# cGAN-Based High Dimensional IMU Sensor Data Generation for Enhanced Human Activity Recognition in Therapeutic Activities

Mohammad Mohammadzadeh[a,b], Ali Ghadami[a,c], Alireza Taheri[a], Saeed Behzadipour[a,b*]

[a] *Department of Mechanical Engineering, Sharif University of Technology, Tehran, Iran*
[b] *Djavad Mowaffaghian Research Center in Neuro-rehabilitation Technologies, Sharif University of Technology, Tehran, Iran*
[c] *Present address: Department of Mechanical Engineering, University of Michigan, Ann Arbor, USA*

**Abstract**

Human activity recognition is a core technology for applications such as rehabilitation, health monitoring, and human-computer interactions. Wearable devices, especially IMU sensors, provide rich features of human movements at a reasonable cost, which can be leveraged in activity recognition. Developing a robust classifier for activity recognition has always been of interest to researchers. One major problem is that there is usually a deficit of training data, which makes developing deep classifiers difficult and sometimes impossible. In this work, a novel GAN network called TheraGAN was developed to generate IMU signals associated with rehabilitation activities. The generated signal comprises data from a 6-channel IMU, i.e., angular velocities and linear accelerations. Also, introducing simple activities simplified the generation process for activities of varying lengths. To evaluate the generated signals, several qualitative and quantitative studies were conducted, including perceptual similarity analysis, comparing manually extracted features to those from real data, visual inspection, and an investigation into how the generated data affects the performance of three deep classifiers trained on the generated and real data. The results showed that the generated signals closely mimicked the real signals, and adding generated data resulted in a significant improvement in the performance of all tested networks. Among the tested networks, the LSTM classifier demonstrated the most significant improvement, achieving a 13.27% boost, effectively addressing the challenge of data scarcity. This shows the validity of the generated data as well as TheraGAN as a tool to build more robust classifiers in case of imbalanced and insufficient data problems.

## 1. Introduction

Today, there are plenty of applications for Human Activity Recognition (HAR) using wearable devices, such as healthcare and rehabilitation [1], Human-computer interaction [2], and fitness [3]. Wearable sensors, including smartphones, smartwatches, smart shoes, etc., are widely used in our lives. Usually, these devices are equipped with Inertial Measurement Units (IMUs). IMU consists of three parts: an accelerometer, a gyroscope, and a magnetometer. Accelerometers and gyroscopes are commonly used in many HAR applications [4]. The main objective of HAR systems is to detect human body motions. This detection can occur through the application of various mathematical models, primarily categorized into statistical and neural network models, which can be used in different applications, including remotely monitoring the physical therapy of patients and detecting neurodegenerative diseases. In recent years, with the improvement of deep learning models, most HAR investigations have moved towards deep networks. Although deep models have some potential advantages, such as automatic feature extraction [5], they have some drawbacks. In these models, the dataset has a decisive effect on the model performance [6]. A rich labeled dataset qualitatively and quantitively helps improve the overall model's performance. In real-world HAR applications, however, collecting data is not easy, for example, in the case of fall detection, medical diagnoses, medical treatment, and anomaly detection [7]. Consequently, collecting a suitable dataset inevitably is time and cost-consuming.

There are some solutions to tackle the problem of data scarcity. One of the solutions is using pre-trained networks as feature extractors. It is possible to extract features from signals using pre-trained networks without the need to train them for a particular dataset, or at least they require only fine-tuning. Ghadami et al. used well-known pre-trained neural networks such as VGG16 [8] to deal with the scarcity of drowsiness data in their works and showed the effectiveness of this method [9]. While transfer learning can mitigate data shortages, pre-trained

---

* Corresponding Author
Email: behzadipour@sharif.edu

  

networks are only useful when their inputs are similar to the data that they were trained for. Another approach is co-training. In co-training, a classifier that is already trained with an existing labeled dataset is used to classify the unlabeled data. Chen et al. trained three classifiers for acceleration, angular velocity, and magnetism [10]. The trained classifiers were used to classify some unlabeled data. If most classifiers have the same prediction on a specific sample, the sample was labeled and added to the dataset. They succeeded in balancing the training samples across their classes using this method. Accumulated error is common in co-training, and the newly labeled data has a limited enhancement in classifier performance.

Active learning is also an effective way to label unlabeled data and solve the problem of data scarcity. In this method, the most informative unlabeled samples are detected and marked by some annotators. Hossain and Roy used active learning by leveraging a new loss function named joint loss function [11]. In this investigation, active learning combined the deep model and the model optimized by the joint loss function. They received help from users as annotators. In active learning, the presence of a witness is necessary, which can be expensive and time-consuming. Also, this method does not use the entire dataset, so the labeled dataset may not be compelling.

Another solution for data scarcity is to enrich the dataset. Data augmentation can enlarge the dataset by synthesizing data. Data augmentation uses small amounts of real data and tries to synthesize fake data similar to real ones. Augmented data can expand input space, prevent overfitting, and improve generalization in deep models [12]. There are several approaches for data augmentation. One of the most popular data augmentation methods is data deformation, which is termed classical data augmentation. The data deformation method is a set of deformations applied to the training dataset to represent unseen raw data. Um et al. used various data augmentation methods for wearable sensor data (IMU sensors) to improve the classification performance for Parkinson's disease monitoring [6]. This investigation applied some position and time-series data augmentation methods, such as rotation, permutation, jittering, scaling, time-warping, and magnitude-warping. In these methods, as the generated data should mimic the real data, it is necessary to ensure that the augmented data maintains the inherent features of an activity's signal and constructs a valid output [13]. This procedure is practical but has some limitations due to the fact that the generated data are only slightly different from the real data. Therefore, it can only cover a close space around the data, leading to a limited increase in classifier accuracy.

Learning-based augmentation methods such as generative adversarial networks (GANs) are another popular tool for synthesizing realistic data. According to some studies, GAN models outperform classical augmentation methods [14]. GAN models are composed of two major parts: generator and discriminator. The generator is a deep model that synthesizes data, and the discriminator is a binary classifier that detects the generated data from real data. The purpose of the generator is to generate realistic data that has similarities with real ones to deceive the discriminator into recognizing them as real. GANs are trained based on the min-max theory in such a way that the discriminator and generator are trained mutually [5]. GAN models are successful in various domains, such as image [15], video [16], language generation [17], and human activity recognition [18]. In wearable-HAR data, there are multi-axial sensor signals. These signals depend on each other, which makes the synthesizing process challenging. For example, in IMU sensors, signals pertain to each other from two aspects, they are time series signals that are made from a specific activity, and the other one is the mechanical relations between velocity and acceleration. Establishing all these relationships in the generated data is necessary to create realistic data.

The first GAN model used in HAR is sensoryGANs [19]. SensoryGAN considers a GAN model for each activity. These models synthesize uniaxial acceleration data of three daily activities: staying, walking, and jogging. These GAN models were customized for each activity. The generators and discriminators were chosen among one-dimensional convolutional neural networks (1D-CNN), Bi-LSTM, and LSTM. Results showed that generated data could be used along with real ones to increase the classifier accuracy. Although sensoryGANs succeeded in creating wearable-HAR data, it is only evaluated on classical classifiers. There is no clue that this model generated data is beneficial for training deep models. Despite the fact that the sensoryGAN model had achieved impressive results for the first impact of GAN in HAR, it could not be modified to synthesize multi-axial sensor data of Wearable-HAR.

However, most datasets collected by IMU sensors have multi-axial channel data, and it is necessary to conform to this data structure to generate meaningful data. In this case, Hu [20] developed a conditional GAN to investigate the imbalanced datasets in human activity. In this study, they tried to synthesize tri-axial acceleration signals of all activities by one GAN model. The conditional input of this model was the embedded label of different activities. The generator and discriminator were built by 1D-CNN and 1D-transposed CNN. It was demonstrated that the synthesized data could solve the problem of imbalanced datasets by obtaining an increased accuracy in an LSTM classifier trained by new balanced datasets. However, this model could only generate data of activities related to simple or repetitive activities. In another work, Shi et al. [18] proposed a GAN model named HAR-Aug GAN to generate tri-axial accelerations of 19 activities. The synthesized data were evaluated on a real test set by training a classifier with synthesized data and testing it with the real test set. The results showed that synthesized data could improve classification by over 10 percent. Nonetheless, the proposed GAN is not applicable for complex activities with a long duration.

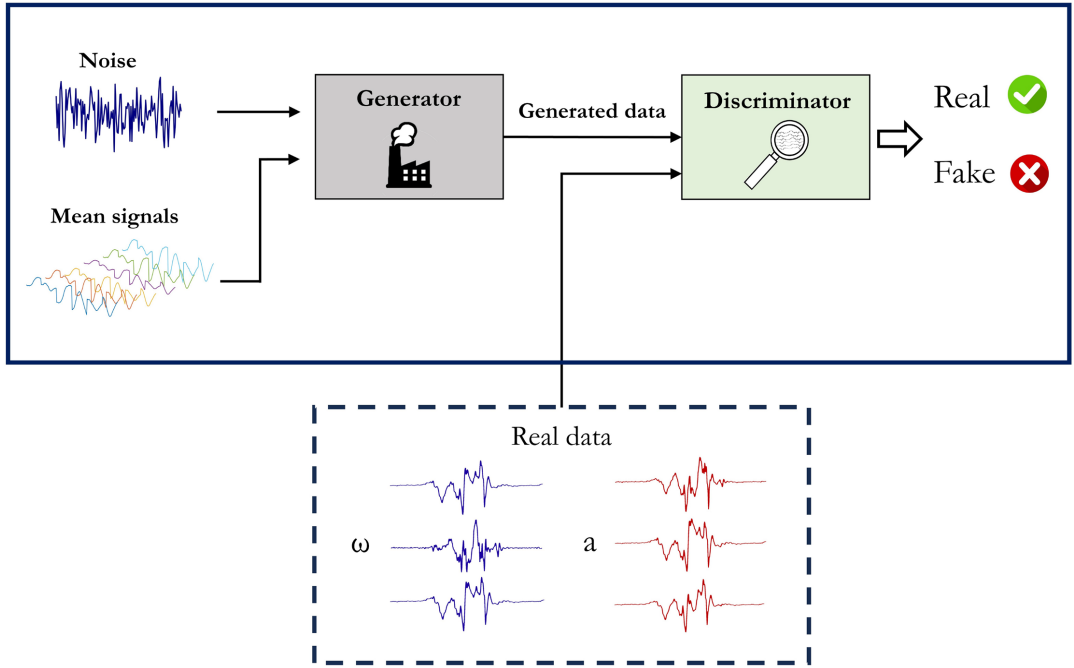

Fig. 1. The overview of the proposed data augmentation method using TheraGAN.

There are several methods to train GAN models. Zhang et al. [21] proposed a GAN model that used a discriminator that classified the labels of intended activities and fake data instead of using a binary classifier as usual discriminators. Besides, the input noise for the generator was not a conventional Gaussian noise. The generator was derived from a part of a variational autoencoder (VAE) structure. Therefore, the generator did not need labels as input, as in regular conditional GANs. Wang et al. used temporal IMU signals simultaneously with their FFT conversions to enhance the discriminator performance in detecting real signals from fake ones [7].

GAN models are confronted with a different problem in the case of sequential data. The ideal augmentation model should continuously synthesize data samples. However, considering that the GAN models are constructed using deep models, the inputs and outputs are limited in size. Also, the sensory data in wearable HAR signals have some spatiotemporal correlations, and it is vital that these correlations also exist in the generated data. In this regard, Wang et al. proposed a new structured conditional GAN model named conditional SensoryGANs [7]. Conditional SensoryGANs managed to generate a longer period of data. As a result, the relationships between different signal parts could be maintained for longer times. In this model, the generator contains MultiScale-MultiDimensional (MSMD) and temporal modules created with CNN and LSTM models. They chose repetitive activities such as walking, which are performed in a short time, and postural activities to assess their model. The duration of these activities did not exceed the model output, or at least one cycle of repetitive movements was done within the range.

Although various works have been done in the field of wearable-HAR data generation using GANs in recent years, nobody has yet reached a model to maintain the long temporal relations between different parts of the generated signal. The nature of human activity is complex. Complex activities can be broken down into unit-level activities [22]. Usually, complex activities take more time than simple activities. So, it is impossible to generate complex activities using the available augmenting models in the literature because of their discrete inherent.

This study investigated the development of TheraGAN, which synthesizes simple therapeutic activities that are the building blocks of complex activities. Our augmentation algorithm was applied to generate multi-axial outputs of four IMU sensors attached to the wrists and thighs. Consequently, 96 GAN models were trained for four IMU sensors to create 24 simple activities extracted from 10 complex activities. We chose more complex activities rather than those found in the literature, i.e., walking and standing, making the augmentation process more challenging. TheraGAN is a conditional GAN, where the generator leverages the average of real signals (Fig.1). It simultaneously uses raw data and frequency spectrogram in the discriminator as inputs to discriminate between real and generated data. The proposed model solves the problem of generating high dimensional and long signals by leveraging custom GAN models and using the concept of simple activities.

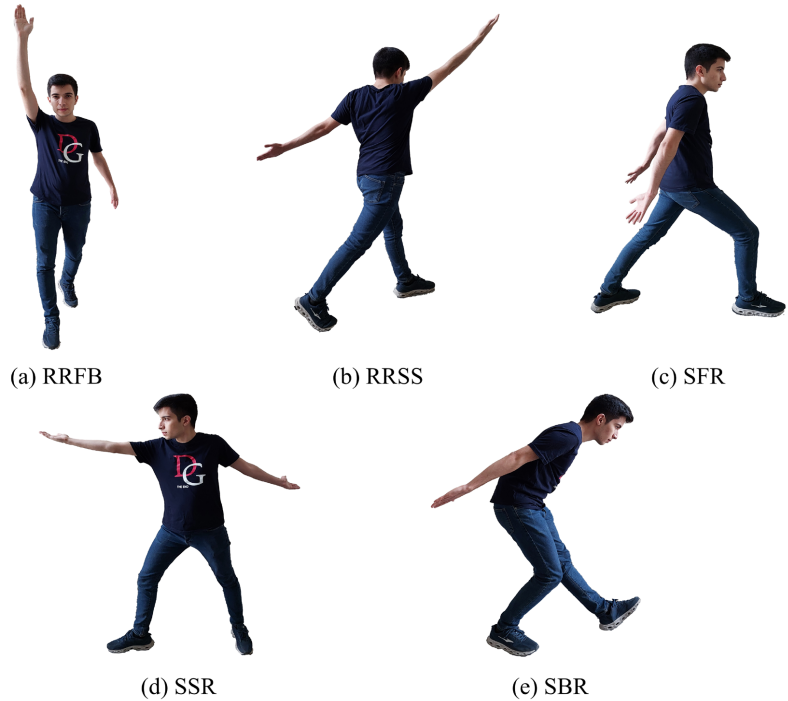

Fig. 2. Selected activities from the dataset. Notably, each activity has two left and right variations. (a) Rock & Reach Forward Backward right (RRFB), (b) Rock and Reach Side to Side right (RRSS), (c) Step Forward & Reach right (SFR), (d) Step to the Side & Reach left (SSR), and (e) Step Backward & Reach left (SBR).

The contributions of this paper can be summarized as follows:

(1) A conditional GAN model named TheraGAN was developed, which could generate multi-axial sensor data (tri-axial accelerometer and tri-axial gyroscope) for long-term complex activities in wearable HAR. The model was employed to generate data from four IMU sensors across ten therapeutic activities.

(2) The quality of the generated data, including the data diversity and similarity, was assessed by various methods, including qualitative and quantitative metrics.

(3) The performance of TheraGAN in generating data was compared with other augmentation methods. This comparison was conducted based on the accuracy of three deep classifiers after training on both real data and generated data obtained from different methods.

The paper's organization is as follows: Section 2 explores the dataset, encompassing its composition and preparatory processes. It also explains the suggested GAN network (TheraGAN) architecture and the training algorithm, accompanied by a detailed overview of the employed evaluation methods on generated data. Moving on to Section 3, it reveals the results and analyses derived from qualitative and quantitative assessments of the generated data. This section further incorporates a comparative analysis of TheraGAN's performance against other augmentation methods. Finally, in Sections 4 and 5, the discussion and conclusions are provided.

## 2. Material and Methods

### 2.1. Dataset

The dataset used in this project was collected at Djavad Mowafaghian Research Centre of Intelligent Neurorehabilitation Technologies. It contains the sensor information of 14 therapeutic activities, which are usually used for exercise therapy in PD patients, collected from 43 healthy individuals aged between 20 and 25 in the laboratory. The significance of this dataset is that each complex activity is divided into several simpler activities, which can be beneficial for both data generation and classification [23]. This dataset was divided into three sections: training, validation, and test. The training dataset contains 31 subjects, while both the validation and test datasets each include 6 subjects. It's important to note that these 12 subjects from the validation and test datasets are not used in the training phase.

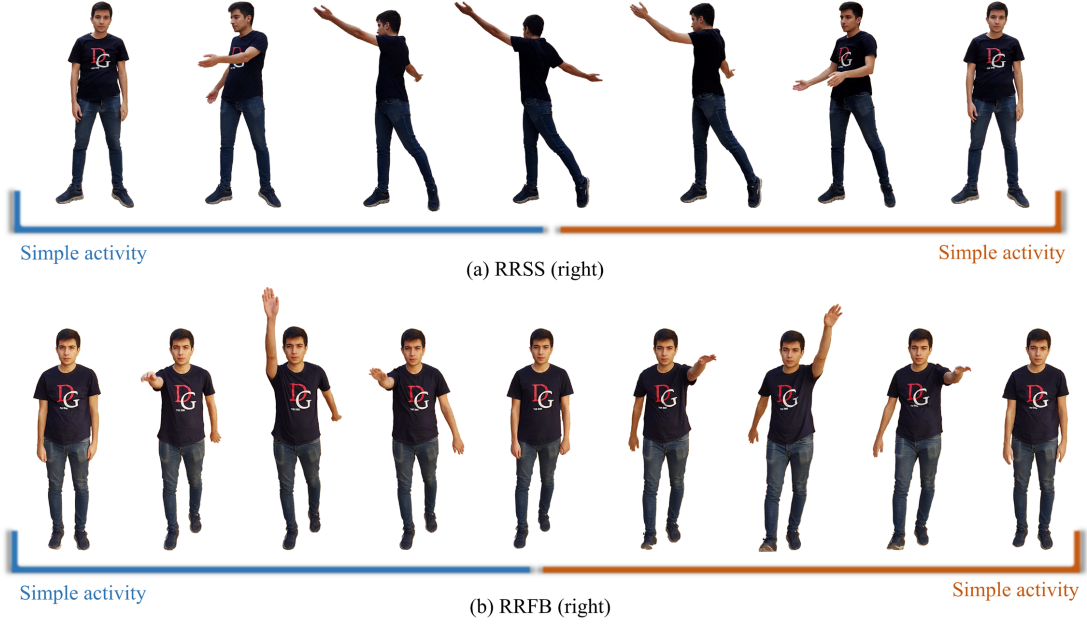

Fig. 3. Detailed overview of the dataset activities with simple activity demonstration for (a) RRSS (right) and (b) RRFB (right)

The data collection system was four inertia sensors mounted on the wrists and thighs of the subjects, which was inspired by another research [24]. The output of each sensor was angular velocities and linear accelerations in the sensor's local coordinate reference at a frequency of 100 Hz. Notably, among these activities, 10 of them for which we had less data were chosen for the rest of the paper. An overview of these chosen therapeutic activities is demonstrated in Fig.2. A more complete demonstration, including simple activity decomposition, is shown for RRSS (right) and RRFB (right) in Fig.3.

*2.2. Data Preprocessing*

The preprocessing in this study comprises two sections: length alignment and normalization. As we aimed to generate simpler divided activities of a complex exercise, the preprocessing was done on the simple activities separately.

Machine learning networks are only able to work with signals of the same length. As a result, the average length of each simple activity was calculated, and the samples with higher length were truncated to the average by randomly removing some frames of the signal. Also, random frames were added to the signal by averaging two consecutive frames for samples whose length was less than the average frames of the signals. Notably, these techniques can be done on the generated signals to change their length, i.e., to alter the speed of the activity, as our model outputs the signals with the fixed length.

In addition to adjusting data lengths, normalizing the data is necessary. The normalization was done on each data channel of every simple activity using Eq.1:

$$x^i_{normalized} = \frac{x^i - \min(x^i)}{\max(x^i) - \min(x^i)}, \quad (1)$$

Where $x^i$ is the $i^{th}$ channel data of the signal.

*2.3. Networks and Training*

*2.3.1. GAN network architecture*
As stated before, ten complex activities, which contain 24 simple activities in total, were considered for the model development. Each activity was presented with data from 4 sensors. Thus, we needed 96 separate GAN networks, where each GAN generated the 6-channel data of a sensor for a unique simple activity. Additionally, the architecture of the GANs was identical for all simple activities and sensors.

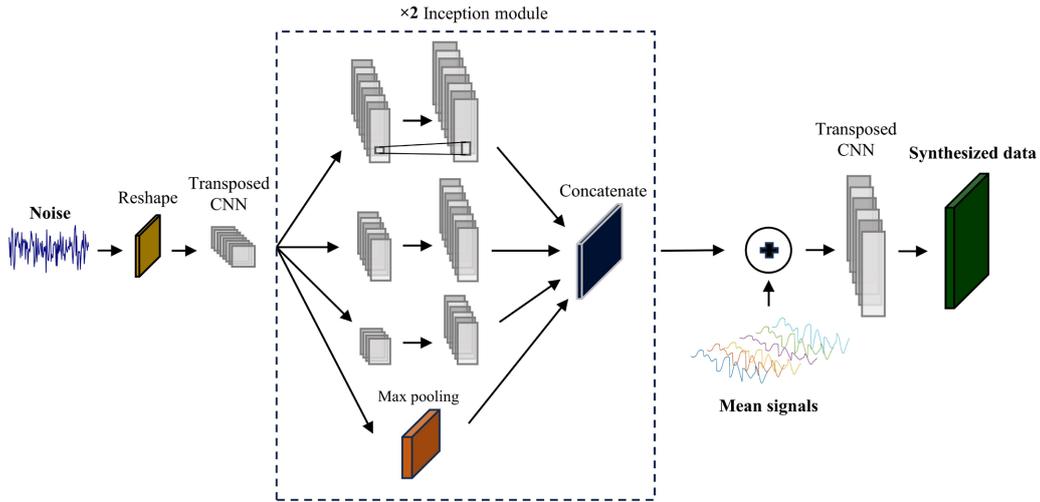

Fig. 4. Proposed architecture for the generator

The generator received a random noise vector of size 128 with a mean value of zero, a standard deviation of one, and the average of the activity signals shape in a same size signal. The proposed network consists of Transposed 1-D Convolutional Neural Networks (1-D T-Conv) arranged in an Inception-like architecture, 1-D Convolutional Neural Networks (1-D CNN), and pooling layers, as shown in Fig.4. The noise signal creates unique initialization for our network, and the averaged signal input guides the network to generate valid outputs. Furthermore, the Inception architecture increases computational effectiveness and helps us extracting features at varying scales. The output is a 6-by-M matrix, which is the concatenation of the data of a single sensor, where M is the average length of the activity, and 6 refers to angular velocities and linear accelerations in 3 directions.

The discriminator is responsible for classifying the input data as fake or real. For this purpose, a dual discriminator composed of a temporal and a frequency classifier was suggested. The temporal classifier has several 1-D convolutional layers for temporal feature extraction, followed by dense layers as a decision maker. This network can perfectly understand the signals' overall behavior but cannot detect oscillations and long-term dependencies. To solve this problem, a second network called frequency classifier was introduced. This network uses the Fast Fourier Transform (FFT) to extract features of the signals in the frequency domain, and then by applying Separable CNNs followed by dense layers, the output is constructed. In addition, Gaussian noise was used during training to avoid overfitting. The combination of the two mentioned classifiers as discriminators (Fig.5) allowed us to consider many aspects of the input signal and construct a robust discriminator.

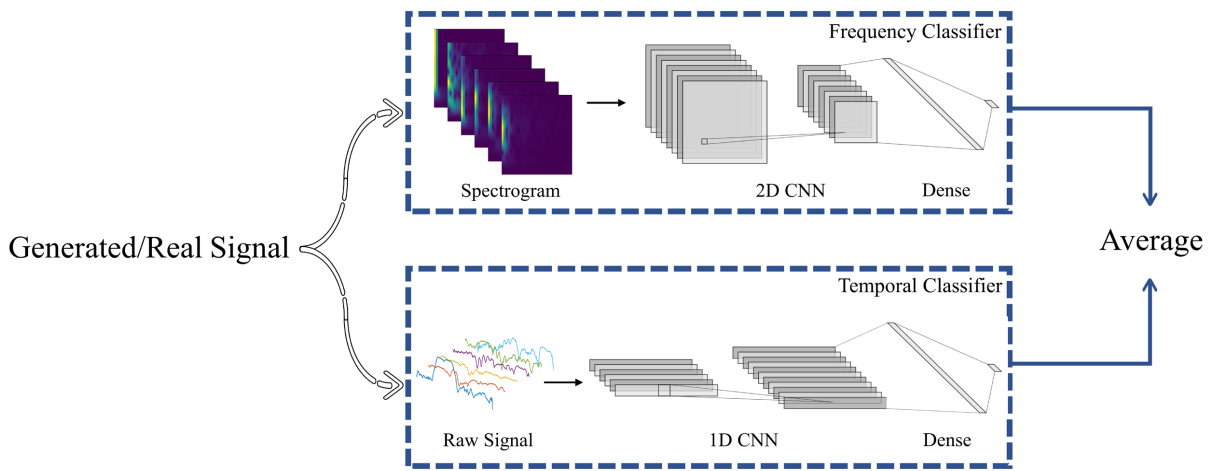

Fig. 5. An overview of the discriminator, which is composed of temporal and frequency classifiers. The final output of the discriminator is the averaged output of the two classifiers.

*2.3.2. Training*

GANs are trained in a min-max game, that is, the discriminator tries to improve itself to classify the generator's outputs as fake, and the generator tries to generate outputs so that the discriminator can no longer recognize it as fake. Equivalently, the network's loss function is:

$$\min_G \max_D V(D,G) = \mathbb{E}_{x \sim p_{data}(x)}[\log D(x)] + \mathbb{E}_{z \sim p_z(z)}\left[\log\left(1 - D\left(G(z, x_{average})\right)\right)\right] \quad (2)$$

The training of our network was a repeating two-step process, such as any other GAN. First, the discriminator was trained. In this stage, random noise vectors with the size of half of the batch size were chosen and fed to the generator to generate data. The data were labeled as fake. Then, random samples of real data were selected as many as in the noise vectors and labeled as real. Then, these signals (real and fake) were fed to the discriminator, and the output was obtained by averaging the output of the two classifiers in the discriminator. In this stage, the generator weights were frozen, and the discriminator was trained by backpropagating the error. The second stage of the training corresponded to the training of the generator. Random vectors of noise to the number of batch size were chosen, and the output of the generator was generated and labeled as real. Then, they were fed to the discriminator as before, and the generator was trained by freezing the discriminator's weights.

The optimum state is determined based on the Nash equilibrium. In Nash equilibrium, the accuracy of the generator reaches around 100 per cent, and the accuracy of the discriminator reaches roughly 50 per cent. It implies that all of the generated data from the generator are labeled as real by the discriminator. Although it seems to be a good criterion, there are some problems and limitations. There are some cases when the Nash equilibrium conditions are reached, but the low accuracy of the discriminator might be because of insufficient training. Therefore, another metric, known as the perceptual similarity metric, is used [25]. In this method, the distance between features of the generated and real images extracted from pre-trained networks are compared by L2-norm

---

**Algorithm 1** TheraGAN training algorithm

**Input:** (1) Gaussian noises $z \sim N(\mu, \sigma)$, (2) real multi-axial data $x_{real} \sim X$;
**Output:** (1) trained simple activity multi-axials generator $G^*$, (2) synthesize data $\tilde{x}$;
Initial $epoch \leftarrow 0$
$count_{discriminator} \leftarrow 0$
$count_{generator} \leftarrow 0$
**repeat**
  **for** i in range $(Batch/2)$ **do**
    Generate fake data $(\tilde{x}^i, \tilde{y}^i)$ using $G(z, X_{ave})$
    randomly sample real data $(x_{real}^i, y_{real}^i)$
  **end for**
  concatenate $(\tilde{x}^i, \tilde{y}^i)$ and $(x_{real}^i, y_{real}^i)$ to get $(x_D^i, y_D^i)$ as input data of discriminator $(D)$
  **repeat**
    train the temporal $(T_D)$ and frequency $(F_D)$ classifiers by $(x_D^i, y_D^i)$
    update the Discriminator $(D)$ parameters $\theta_D$ using gradients of loss:
    $\nabla_{\theta_D} \frac{1}{t} \sum_{i=1}^{t} \left[\log D(\tilde{x}^i) + \log(1 - D(x_{real}^i))\right]$
    $count_{discriminator} + +;$
  **until** ($T_D$ and $F_D$ losses < 0.1 or $count_{discriminator}$ > 20)
  **for** i in range $Batch$ **do**
    Generate fake data $(\tilde{x}^i, \tilde{y}^i)$ using $G(z^i, X_{ave})$
    Sample real data $(x_{real}^i, y_{real}^i)$
  **end for**
  Define discriminator output as:
  $D = (T_D + F_D)/2$
  **repeat**
    update the Generator $(G)$ parameters $\theta_G$ using gradients of loss:
    $\nabla_{\theta_G} \frac{1}{batch} \sum_{i=1}^{batch} \left[\log D\left(G(z^i, X_{ave})\right)\right]$
    $count_{generator} + +;$
  **until** (generator's loss < 0.12 or $count_{generator}$ > 50)
  Calculate the perceptual similarity distance $S_d$ between generated data using trained generator and real data
  $epoch + +;$
**until** (epoch > 90 or $S_d$ < 0.1)

Fig. 6. TheraGAN training algorithm

distance. If the distance is less than a specific value, one can infer that the generator is doing its job well. We used this metric by comparing the L2-norm distance of the features of the spectrogram of the generated signals with real ones. VGG16 was chosen as a feature extractor in this work, and the L2-norm threshold for stopping the training was chosen to be 0.1. The detailed training process of TheraGAN is depicted in Fig.6.

*2.4. Evaluation Methods*

The usability of the generated data must be assessed to ensure that the main goal of this study, which is generating data similar to the real data to enhance the dataset quality and quantity, is achieved. To justify this aim, we resorted to two different kinds of assessments.

*2.4.1. Qualitative assessment*

The quality of generated data can be evaluated in three ways:
(1) **Diversity:** The diversity of the generated data is one of the metrics that can show how well we performed in data augmentation. GAN models are susceptible to creating repetitive data (mode collapse issue). In this case, it is crucial for the GAN models to be designed and trained in a way that they do not stick to limited synthetic data. Several generated signals should be checked visually to ensure the devised GAN model outcome maintains favorable creativity.
(2) **Similarity:** The generated data must be analogous to real data. This purpose can be fulfilled by using visual inspection. In the data acquired from particular human activities such as rehabilitative ones, we can see some unique temporal and spatial properties in wearable sensor signals. These traits can be detected visually, making the comparison between the generated and real data possible. Also, we had previously made sure that the generated data were similar to real ones using perceptual similarity checks during training. A comparison between the heat map of the generated and real signals was made to address the diversity and similarity.
(3) **Visualization with Uniform Manifold Approximation and Projection (UMAP):** It is possible to go further by extracting some meaningful features such as temporal and spatial features from the IMU signals and using UMAP to visualize them in two dimensions. This visualization was applied to both the generated and real data. If, in this two-dimensional visualization, the generated and real signals were closely situated, and the generated data filled the gaps near the real ones, we concluded that these data exhibited similarity.

*2.4.2. Quantitative assessment*

In this assessment, three well-known deep machine-learning models, LSTM, CNN, and transformer networks, were trained in the presence of the generated data alongside real ones. Then, the performance of these models was compared with their previous performance when they were only trained with the prior unenhanced dataset (only the real data) to see if any improvement had occurred.

Among our models, the CNN classifier comprised a 2D-Inception module followed by dense layers, the LSTM network was a three-layer LSTM followed by dense layers, and finally, the transformer classifier was a single transformer encoder. Transformers were chosen as they are cutting-edge sequence modeling networks and solve the problem of remembering long dependencies and parallelization by introducing the attention module. The inputs of the sequential models, LSTM cells and transformer networks embedding input were 450 vectors with the size of 24×1, which contained the angular velocities and accelerations signal information of 4 sensors and a matrix with a size of 24×450 was fed to our CNN classifier. The size of the input data for all models was fixed to

Table 1. Hyperparameters of the LSTM model

| Layer | # of units | Activation |
|---|---|---|
| Input | | |
| LSTM | 256 | tanh |
| LSTM | 128 | tanh |
| LSTM | 64 | tanh |
| Fully connected | 64 | Leaky ReLU |
| Fully connected | 32 | Leaky ReLU |
| Fully connected | 32 | Leaky ReLU |
| Fully connected | 10 | softmax |
| | | |
| Optimizer: | Adam (lr=0.001) | |

Table 2. Hyperparameters of the CNN model

| Layer | # of units | Kernel size | Activation |
|---|---|---|---|
| Input | | | |
| Inception module | (128,64,32) | (16,8,4) | Leaky ReLU |
| CNN 2D | 32 | 4 | Leaky ReLU |
| CNN 2D | 1 | 16 | Leaky ReLU |
| Fully connected | 32 | | Leaky ReLU |
| Fully connected | 32 | | Leaky ReLU |
| Fully connected | 32 | | Leaky ReLU |
| Fully connected | 10 | | softmax |
| | | | |
| Optimizer: | Adam (lr=0.001) | | |

Table 3. Hyperparameters of the transformer model

|   | Layer | # of units | Activation |
|---|---|---|---|
| 3x | Input<br>attention module<br>Fully connected<br>Fully connected | <br><br>(128,128,24)<br>10 | <br><br>tanh<br>softmax |
|   | Optimizer: | Adam (lr=0.001) |   |

450 by sliding a window over a complex activity with a stride of 225. i.e., a complex activity of length of 900 provided three data samples. This was to resemble the real time usage of these classifiers. The hyper parameters and the structure of the implemented models are demonstrated in Table.1, Table.2 and Table.3.

## 3. Results and analysis

As discussed earlier, 96 separate GANs were trained for 10 different activities, and 19592 data samples were used for training. After training the networks, the augmented complex activities were constructed by concatenating the generated data of the simple activities generated from the GAN's generators. This way, one can construct an infinite number of complex activities by concatenating the generated simple activities sequentially. As a result, each generated complex activity is unique. It should be noted that the end and start points of simple activities might not align perfectly. Hence, the connection region for these activities was slightly altered (implementing interpolation in the interval of signals) to maintain continuity.

*3.1. Qualitative evaluation result*

For this section, several generated complex activity signals were inspected, and the comparison result shows that the generated data could perfectly follow the trend of the real data for every signal channel. This similarity between generated and real data without any temporal limitation is remarkable for TheraGAN compared to other existing GAN models in the literature. To demonstrate a similar distribution of generated data to real data, the heatmaps of one simple activity for one of the sensors were illustrated in Fig.7. Each heatmap was constructed from 50 samples. As evident in Fig.5, the distribution of signals is similar to each other, with only minor variations, and the generator can capture the diversity in the generated signals.

An example of the synthesized data generated by TheraGAN and its closest real equivalent (determined by perceptual similarity metrics) is illustrated in Fig.8. As shown, the complex activity is the concatenation of the generated signals for each channel.

In addition, for each of the 24 IMU sensor output channels, 62 different features were extracted. These features were recommended by [26-29]. and include time, frequency, and time-frequency features. Using a random forest model, 40 more optimal features for classification were selected. Thereafter, using UMAP dimensionality reduction technique, the dimension of these features decreased to two dimensions. Scatter charts illustrating the two principal features extracted using UMAP for each activity are shown in Fig. 9. Based on these charts, we can conclude that the generated and real data for all activities exhibit similar features, suggesting that the generated data in conjunction with its similarity to real data, can complement the real data.

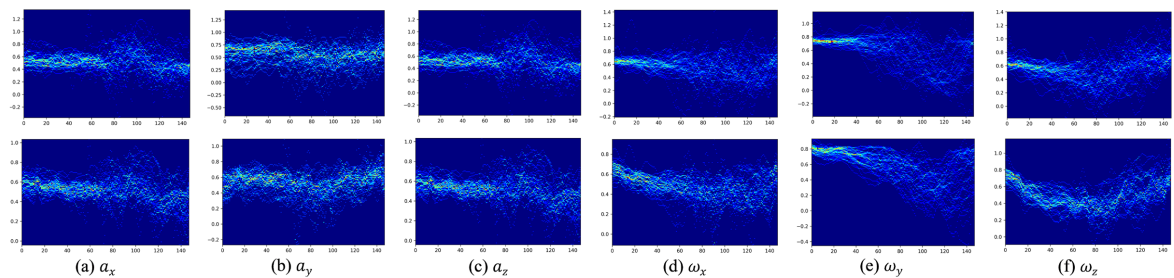

(a) $a_x$   (b) $a_y$   (c) $a_z$   (d) $\omega_x$   (e) $\omega_y$   (f) $\omega_z$

Fig. 7. Heatmap depicting real and generated acceleration and angular velocity signals during the second simple activity in SFR (Left) complex activity. The images in the first row correspond to real data, while the images in the second row represent generated data.

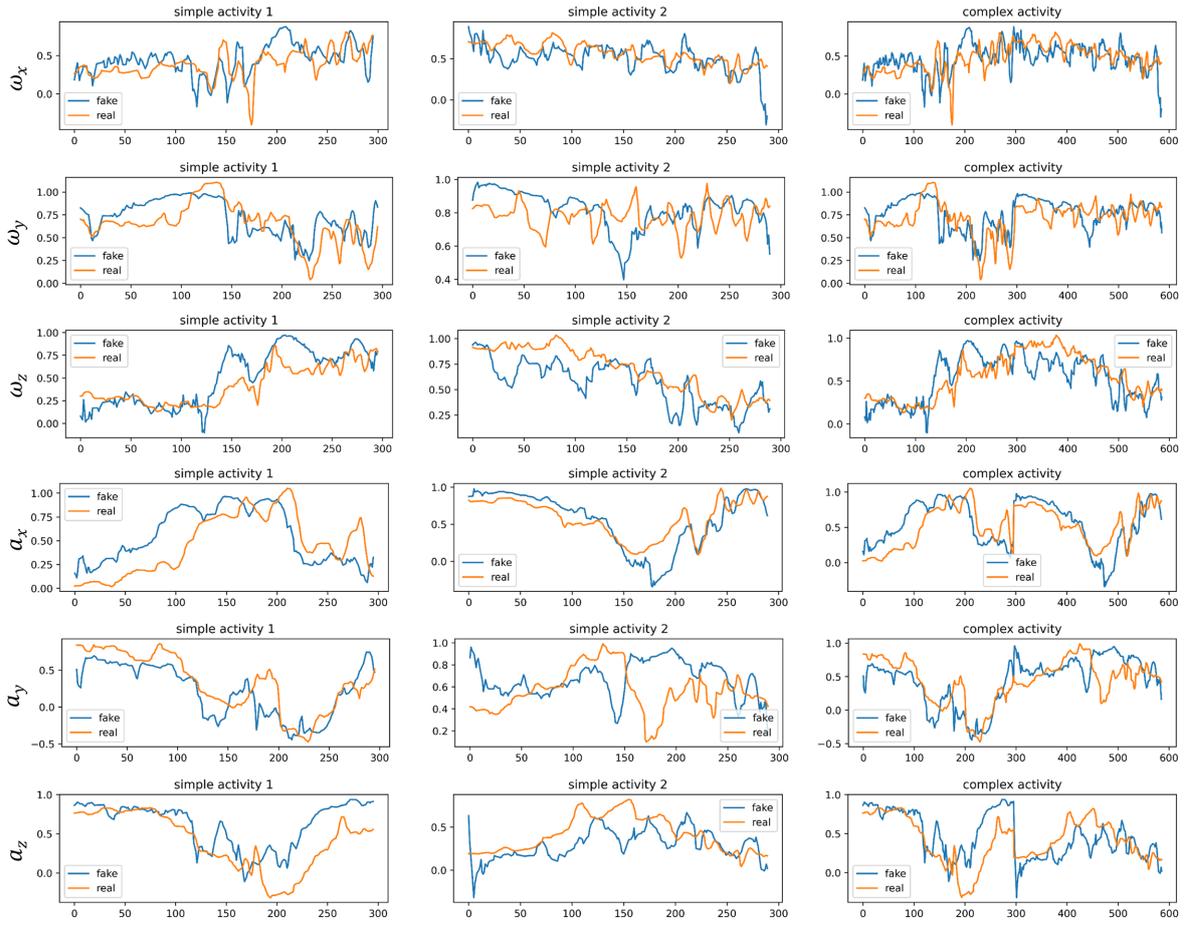

Fig. 8. An example of the generated simple activities of one sensor and the concatenated full-length complex activity (SFR (Left)).

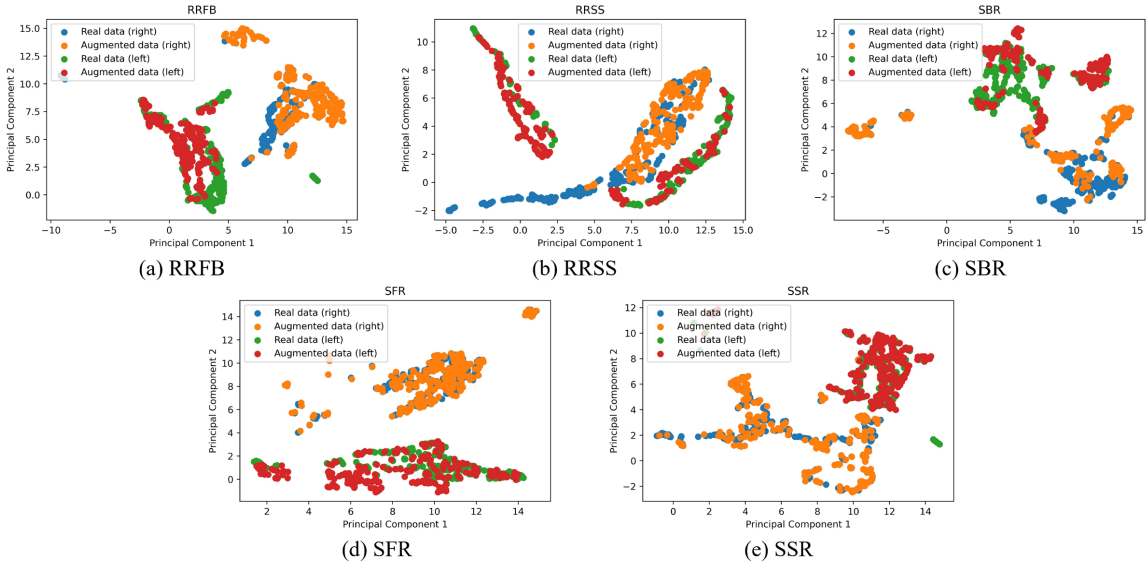

(a) RRFB  (b) RRSS  (c) SBR
(d) SFR  (e) SSR

Fig. 9. UMAP visualization of all activities (the right and left of rehabilitation activity are illustrated in one chart, and only 250 representative samples are considered for clarity.)

*3.2. Quantitative evaluation result*

While the perceptual similarity criterion guides the termination of TheraGAN model training by ensuring the generation of data closely resembling real data, the primary objective of synthesizing data is to enrich the dataset. This enrichment, in turn, contributes to enhancing the model's performance after training with the augmented dataset. Therefore, a more sensible and reliable approach for evaluating the effectiveness of the generated data is to assess their impact on enhancing classifiers' performance through quantitative measures. In this case, the performance of the proposed classifier is related to its capability to improve the classification of various complex activities through the integration of both generated and real data during training. The classification process involves employing a sliding window comprising 24 feature vectors, encompassing angular velocities and accelerations captured by four IMU sensors as inputs to the classifier. This task is more complicated than classifying the complex activity by taking the whole activity's signal as input, and it is more suitable for real-time applications.

As discussed, three different classifiers were used to evaluate the generated data. These classifiers were trained in two different scenarios to assess the generated data. First, the classifiers were trained only by real data. Then, different amounts of the generated data were added to the real data for retraining. After retraining, the networks were tested by data from subjects that were excluded from the training, test dataset. The results of these improvements based on the amount of the added data are shown in Fig.10. These results are reported for 25 runs of the network, and the average is depicted as well. The maximum increase in the average f1-score belongs to the LSTM classifier (13.27%), then the transformer classifier (12.1%), and finally the CNN (11.8%). Also, by performing independent t-test between the results (f1-scores of the networks with and without the generated data),

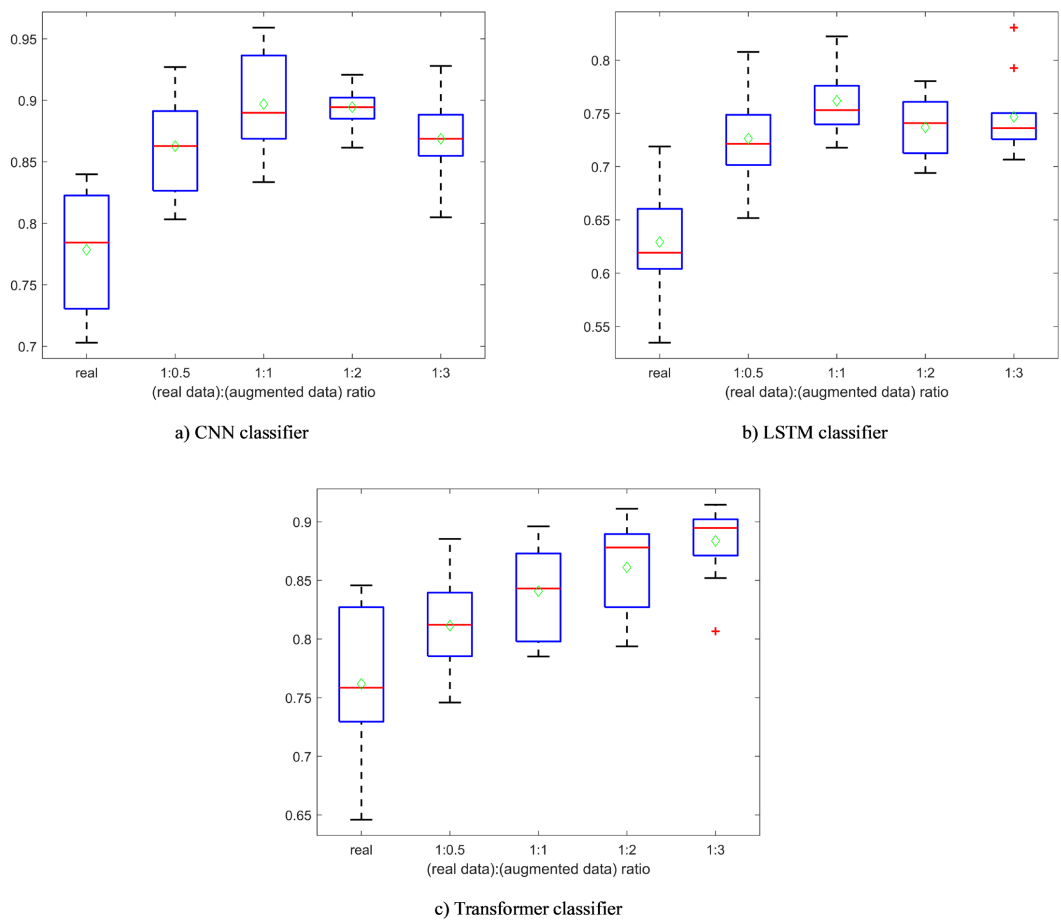

Fig. 10. The performance of the classifiers by adding the generated data in training. The results are provided for ten runs, and the test data is chosen so the networks do not see that during training (different subjects).

*Preprint submitted to Elsevier* 11

Table 4. T-test analysis of the F1-socres of the networks with and without the synthetic data.

| Network | Real:Augmented Ratio | Mean (SD) | | T-Value | P-Value |
|---|---|---|---|---|---|
| | | Only real | Real+Augmented | | |
| LSTM | 1:0.5 | 0.629 (0.049) | 0.726 (0.043) | -4.44 | $2.2 \times 10^{-4}$ |
| | 1:1 | | 0.761 (0.032) | -6.76 | $6.4 \times 10^{-6}$ |
| | 1:2 | | 0.736 (0.028) | -5.70 | $3.1 \times 10^{-7}$ |
| | 1:3 | | 0.751 (0.034) | -6.09 | $4.6 \times 10^{-5}$ |
| CNN | 1:0.5 | 0.778 (0.046) | 0.862 (0.037) | -4.49 | $7.0 \times 10^{-4}$ |
| | 1:1 | | 0.897 (0.041) | -6.05 | $6.3 \times 10^{-5}$ |
| | 1:2 | | 0.894 (0.016) | -7.50 | $8.9 \times 10^{-5}$ |
| | 1:3 | | 0.869 (0.030) | -5.18 | $1.3 \times 10^{-3}$ |
| Transformer | 1:0.5 | 0.762 (0.059) | 0.811 (0.039) | -2.07 | 0.06 |
| | 1:1 | | 0.841 (0.037) | -3.36 | $3.4 \times 10^{-3}$ |
| | 1:2 | | 0.861 (0.041) | -4.11 | $6.6 \times 10^{-4}$ |
| | 1:3 | | 0.884 (0.031) | -5.44 | $3.6 \times 10^{-5}$ |

it can be found that by adding the generated data to the networks, almost always there will be a significant improvement in the performance (p-value less than 0.05). These results are also available in Table.4.

*3.3. Performance of TheraGAN against other augmentation methods*

TheraGAN, to the best of our knowledge, is the first GAN model that uses a segmented wearable sensor dataset and manages to generate long-term time-series IMU signals. Additionally, TheraGAN can simultaneously generate the IMU sensor's six axes' simultaneously. Therefore, there is no GAN model like TheraGAN to compare with. In this case, one kind of augmentation approach that is useful for long-term wearable sensor signals is adding minor changes to the data in a way that does not change the data label. These augmentation methods can be counted as adding Gaussian noise, rotation, scaling, magnitude, and time warping, which are suggested in [6, 30]. The configuration of these augmentation methods precisely adheres to the details outlined in these articles. We trained all deep classifiers, including CNN, LSTM, and Transformer encoders, using this augmented data at a one-to-one ratio with the real data, conducting ten training sessions. The average performance of these classifiers, particularly the f1-score, is illustrated in Table 5. Based on the results, TheraGAN achieves the highest accuracy between different methods in all of the tested networks.

## 4. Discussion

In this study, we introduced a conditional GAN model, TheraGAN, designed to generate signals corresponding to four IMU sensors attached to the wrists and thighs, capturing human movements during therapeutic exercises. The generated data serves two purposes: balancing and enlarging the dataset, aiming to enhance the accuracy of deep activity classifiers when trained with it. The results of this augmentation model demonstrate several notable achievements.

First, TheraGAN has the capability to synthesize six axial signals (tri-axial acceleration and tri-axial angular velocities) obtained from each IMU sensor attached to different body parts, such as wrists and thighs. Therefore, the augmented data dimension equals the real ones, and there is no dimension reduction. This property facilitates the use of augmented data along with actual data.

Table 5. F1-Scores of deep classifiers trained with various augmentation methods and real data at one-to-one ratio. Mean performance and standard deviations are presented.

| | TheraGAN | Noise | scaling | rotation | Mag-warping | Time-warping |
|---|---|---|---|---|---|---|
| CNN classifier | **89.7** (0.04) | 86 (0.06) | 80.2 (0.1) | 62.4 (0.09) | 71.3 (0.04) | 68.8 (0.07) |
| LSTM classifier | **76.2** (0.03) | 73.8 (0.03) | 72.2 (0.04) | 69.2 (0.04) | 75.6 (0.04) | 74.7 (0.04) |
| Transformer classifier | **84.1** (0.04) | 82.3 (0.03) | 82.9 (0.04) | 81.3 (0.03) | 81.3 (0.04) | 80.9 (0.02) |

Secondly, TheraGAN has the ability to generate continuous signals by creating the signals of simple activities and then combining them to construct complex activity signals. This unique feature addresses the challenges associated with discretizing long-term data for use in GAN models, allowing TheraGAN to synthesize entire complex activity signals without limitations on activity length. This characteristic proves beneficial for real-world applications.

Thirdly, synthesized and real data are used as training data in conventional and cutting-edge deep classifiers. The results show that the synthesized data can be used along with real ones and improve the classifier's accuracy, at least for the activities of this work. In this context, TheraGAN produced a more realistic and elongated time series that outperforms available models in generating realistic IMU data. Also, the results of the transformer classifier showed that adding generated data constantly improves the network's performance. This may be due to the fact that transformer networks need a lot of data for training, and when adequate data is available, very high accuracies can be achieved.

Finally, in a comparison of three classifiers trained with the data generated from TheraGAN and other augmentation methods, it is evident that the performance of classifiers in the presence of data generated by TheraGAN is superior to that achieved when trained with alternative augmentation approaches. This achievement can be attributed to two primary reasons. Firstly, within the GAN model's structure, an inspector evaluates the generated data and prevents the generation of data that deviates significantly from real examples. Secondly, data generated by GAN models has the ability to encompass a broader domain of real data environments compared to data generated through augmentation methods.

Although the suggested method solved the problem of the imbalanced dataset in wearable HAR, there are some limitations that should be considered. First, this method requires especially collected datasets, where simple activities are labeled and annotated manually. This adds to the complexity of data collection and can be time-consuming. Second, data for each simple activity should be generated using a number of independent GAN networks, with each network corresponding to one of the IMU sensors used for data collection. In this case, many GAN networks should be trained, which might not be a good idea when the number of simple activities exceeds a certain number due to the need for high computational power and training time. To overcome the identified limitations, further investigations are necessary to design a GAN model that does not rely on labeled datasets for the constituent simple activities within complex ones. Such an advancement would enhance the autonomy of the data generation model. Another problem is the significant number of required GAN models for generating data. In this case, a new approach should be suggested to reduce the number of necessary GAN models to create simple activities. This approach can minimize training time and contribute to overall model efficiency.

## 5. Conclusions

In this paper, we proposed a conditional generative model named TheraGAN to create augmented therapeutic data. TheraGAN can synthesize the six signals related to tri-axial accelerations and tri-axial angular velocities obtained from an IMU sensor. The generated signals for the simple activities were concatenated to create complex ones. TheraGAN's generator has two inputs: Gaussian noise and averages of the signals from which the conditional trait of our GAN model originates. The discriminator in our model consists of two classifiers, temporal and frequency CNN-based binary classifiers. These classifiers complement each other to perceive the temporal and spatial characteristics of the IMU data and make the discriminator appropriate to distinguish real and generated data. TheraGAN can surpass the present articles in synthesizing time series in two aspects. The first is its ability to generate elongated signals related to complex activities without any limitation on the activity duration. The second advantage is its potential capability to generate all IMU sensor signals containing acceleration and angular velocities in their triaxial local coordinate system. The mentioned abilities make TheraGAN more progressive augmentation model for synthesizing IMU continuous data in complex activities. Experiments show that generated data by our model can improve the performance of deep classifiers in HAR clinical Parkinson's exercises. Although TheraGAN enabled the generation of multi-axial continuous data, it still has some deficiencies that must be addressed. The main challenge is its prerequisite to a labeled dataset in which simple activity labels are specified. This means that meaningful, simple activities that make complex activities must be determined before collecting the dataset. The second deficiency is the number of GAN models that are necessary to create every simple activity for each sensor using a large number of GAN models prolonged the training and necessitated powerful computer hardware.

# References

[1] H. B. Kim *et al.*, "Wrist sensor-based tremor severity quantification in Parkinson's disease using convolutional neural network," *Computers in biology and medicine,* vol. 95, pp. 140-146, 2018.

[2] H. Zhao, Y. Ma, S. Wang, A. Watson, and G. Zhou, "MobiGesture: Mobility-aware hand gesture recognition for healthcare," *Smart Health,* vol. 9, pp. 129-143, 2018.

[3] D. Morris, T. S. Saponas, A. Guillory, and I. Kelner, "RecoFit: using a wearable sensor to find, recognize, and count repetitive exercises," in *Proceedings of the SIGCHI Conference on Human Factors in Computing Systems*, 2014, pp. 3225-3234.

[4] S. Zhang *et al.*, "Deep learning in human activity recognition with wearable sensors: A review on advances," *Sensors,* vol. 22, no. 4, p. 1476, 2022.

[5] K. Chen, D. Zhang, L. Yao, B. Guo, Z. Yu, and Y. Liu, "Deep learning for sensor-based human activity recognition: Overview, challenges, and opportunities," *ACM Computing Surveys (CSUR),* vol. 54, no. 4, pp. 1-40, 2021.

[6] T. T. Um *et al.*, "Data augmentation of wearable sensor data for parkinson's disease monitoring using convolutional neural networks," in *Proceedings of the 19th ACM international conference on multimodal interaction*, 2017, pp. 216-220.

[7] J. Wang, Y. Chen, and Y. Gu, "A wearable-HAR oriented sensory data generation method based on spatio-temporal reinforced conditional GANs," *Neurocomputing,* vol. 493, pp. 548-567, 2022.

[8] K. Simonyan and A. Zisserman, "Very deep convolutional networks for large-scale image recognition," *arXiv preprint arXiv:1409.1556,* 2014.

[9] A. Ghadami, M. Mohammadzadeh, M. Taghimohammadi, and A. Taheri, "Automated Driver Drowsiness Detection from Single-Channel EEG Signals Using Convolutional Neural Networks and Transfer Learning," in *2022 IEEE 25th International Conference on Intelligent Transportation Systems (ITSC)*, 2022: IEEE, pp. 4068-4073.

[10] K. Chen, L. Yao, D. Zhang, X. Wang, X. Chang, and F. Nie, "A semisupervised recurrent convolutional attention model for human activity recognition," *IEEE transactions on neural networks and learning systems,* vol. 31, no. 5, pp. 1747-1756, 2019.

[11] H. S. Hossain and N. Roy, "Active deep learning for activity recognition with context aware annotator selection," in *Proceedings of the 25th ACM SIGKDD International Conference on Knowledge Discovery & Data Mining*, 2019, pp. 1862-1870.

[12] C. Shorten and T. M. Khoshgoftaar, "A survey on image data augmentation for deep learning," *Journal of big data,* vol. 6, no. 1, pp. 1-48, 2019.

[13] O. Steven Eyobu and D. S. Han, "Feature representation and data augmentation for human activity classification based on wearable IMU sensor data using a deep LSTM neural network," *Sensors,* vol. 18, no. 9, p. 2892, 2018.

[14] M. Frid-Adar, I. Diamant, E. Klang, M. Amitai, J. Goldberger, and H. Greenspan, "GAN-based synthetic medical image augmentation for increased CNN performance in liver lesion classification," *Neurocomputing,* vol. 321, pp. 321-331, 2018.

[15] T. Karras, S. Laine, M. Aittala, J. Hellsten, J. Lehtinen, and T. Aila, "Analyzing and improving the image quality of stylegan," in *Proceedings of the IEEE/CVF conference on computer vision and pattern recognition*, 2020, pp. 8110-8119.

[16] M. Chu, Y. Xie, J. Mayer, L. Leal-Taixé, and N. Thuerey, "Learning temporal coherence via self-supervision for GAN-based video generation," *ACM Transactions on Graphics (TOG),* vol. 39, no. 4, pp. 75: 1-75: 13, 2020.

[17] M. A. Haidar and M. Rezagholizadeh, "Textkd-gan: Text generation using knowledge distillation and generative adversarial networks," in *Advances in Artificial Intelligence: 32nd Canadian Conference on Artificial Intelligence, Canadian AI 2019, Kingston, ON, Canada, May 28–31, 2019, Proceedings 32*, 2019: Springer, pp. 107-118.

[18] J. Shi, D. Zuo, and Z. Zhang, "A GAN‐based data augmentation method for human activity recognition via the caching ability," *Internet technology letters,* vol. 4, no. 5, p. e257, 2021.

[19] J. Wang, Y. Chen, Y. Gu, Y. Xiao, and H. Pan, "SensoryGANs: an effective generative adversarial framework for sensor-based human activity recognition," in *2018 International Joint Conference on Neural Networks (IJCNN)*, 2018: IEEE, pp. 1-8.

[20] Y. Hu, "BSDGAN: Balancing Sensor Data Generative Adversarial Networks for Human Activity Recognition," in *2023 International Joint Conference on Neural Networks (IJCNN)*, 2023: IEEE, pp. 1-8.

[21] X. Zhang, L. Yao, and F. Yuan, "Adversarial variational embedding for robust semi-supervised learning," in *Proceedings of the 25th ACM SIGKDD International Conference on Knowledge Discovery & Data Mining*, 2019, pp. 139-147.